\pdfoutput=1

\documentclass[11pt]{article}

\usepackage[]{ACL2023}

\usepackage{times}
\usepackage{latexsym}

\usepackage[T1]{fontenc}

\usepackage[utf8]{inputenc}

\usepackage{microtype}

\usepackage{inconsolata}

\usepackage{threeparttable}
\usepackage{multirow}
\usepackage{booktabs}
\usepackage{makecell}
\usepackage{amsmath}
\usepackage{amssymb}
\usepackage{graphicx}
\graphicspath{ {./images/} }
\usepackage{svg}
\svgpath{ {./images/} }
\usepackage{mathtools}
\usepackage{xcolor}
\usepackage{enumitem}
\usepackage{qtree}
\usepackage{tikz-qtree,tikz-qtree-compat}
\usepackage{caption}
\usepackage{subcaption}
\usepackage{bbm}
\usepackage{bm}
\usepackage{hypcap}
\usepackage[normalem]{ulem}

\definecolor{bg}{RGB}{70, 129, 70}

\def\Snospace~{\S{}}

\let\orgautoref\autoref
\providecommand{\Autorefs}
        {\def\figureautorefname{Figs.}%
         \def\subfigureautorefname{Figs.}%
         \def\Itemautorefname{Items}%
         \def\tableautorefname{Tables}%
         \orgautoref}

\renewcommand{\autoref}
        {\def\figureautorefname{Fig.}%
         \def\subfigureautorefname{Fig.}%
         \def\sectionautorefname{\Snospace}%
         \def\subsectionautorefname{\Snospace}%
         \def\subsubsectionautorefname{\Snospace}%
         \def\Itemautorefname{item}%
         \def\tableautorefname{Table}%
         \orgautoref}

\DeclareMathOperator*{\argmax}{arg\,max}

\newcommand\tabcaption{\def\@captype{table}\caption}
\newcommand\figcaption{\def\@captype{figure}\caption}

\title{Distributional Inclusion Hypothesis and Quantifications: Probing for Hypernymy in Functional Distributional Semantics}

\author{Chun Hei Lo$^1$ \qquad  Wai Lam$^1$ \qquad Hong Cheng$^1$ \qquad Guy Emerson$^2$ \\
  $^1$The Chinese University of Hong Kong \qquad $^2$University of Cambridge \\
  \texttt{\{chlo, wlam, hcheng\}@se.cuhk.edu.hk} \qquad 
 \texttt{gete2@cam.ac.uk} \\
  }

\begin{document}
\maketitle
\begin{abstract}

Functional Distributional Semantics (FDS) models the meaning of words by truth-conditional functions.
This provides a natural representation for hypernymy but no guarantee that it can be learnt when FDS models are trained on a corpus.
In this paper, we probe into FDS models and study the representations learnt, drawing connections between quantifications, the Distributional Inclusion Hypothesis (DIH), and the variational-autoencoding objective of FDS model training.
Using synthetic data sets,
we reveal that FDS models learn hypernymy on a restricted class of corpus that strictly follows the DIH.
We further introduce a training objective that both enables hypernymy learning under the reverse of the DIH and 
improves hypernymy detection from real corpora.


\end{abstract}

\section{Introduction}
\label{sec:intro}

Functional Distributional Semantics \citep[FDS; ][]{emerson-copestake-2016-functional,emerson2018} suggests that the meaning of a word can be modelled as a truth-conditional function, whose parameters can be learnt using the distributional information in a corpus \citep{emerson-2020-autoencoding, lo-etal-2023-functional}.
Aligning with truth-conditional semantics, functional representations of words are logically more rigorous than vectors (e.g., \citealp{mikolov2013vector, pennington2014glove, levy-goldberg-2014-dependency, czarnowska-etal-2019-words}) and distributions (e.g., \citealp{vilnis2014gauss}, \citealp{brazinskas-etal-2018-embedding}) as concepts are separated from their referents (for a discussion, see: \citealp{emerson-2020-goals,emerson2023}).
On top of its theoretical favour, \citet{lo-etal-2023-functional} also demonstrate FDS models in action and show that they are very competitive in the semantic tasks of semantic composition and verb disambiguation.

Hypernymy is also known as lexical entailment.
It is formally defined as the subsumption of extensions between two word senses, which can be modelled with truth-conditional functions.
Although FDS provides the basis for embedding hypernymy, it is not obvious whether hypernymy can be learnt by training an FDS model on a corpus, and if so,
what kind of corpus the model can successfully learn from.

To acquire hypernymy automatically from a corpus, one way is through the use of distributional information.
In this class of methods, hypernymy is learnt in an unsupervised manner given certain hypotheses about the distributional properties of the corpus.
One such hypothesis is the Distributional Inclusion Hypothesis \citep[DIH; ][]{weeds-etal-2004-characterising,  geffet-dagan-2005-distributional}, which relates lexical entailment of words to the subsumption of the typical contexts they appear with in a corpus.



 
In this paper, we relate the DIH to quantifications, and to FDS.
In \autoref{sec:qtfc_dih}, we first revisit the DIH and highlight
that, while existential quantifications support the DIH,
universal quantifications reverse it.
In~\autoref{sec:fds}, we introduce FDS, discuss how hypernymy can be represented in FDS, and explain how FDS can handle quantifications, thus allowing hypernymy learning under both the DIH and the reverse of it.
Finally, we present experimental results of applying FDS models to both synthetic and real data sets in \autoref{sec:exp_synth} and \autoref{sec:exp_real} respectively.

\section{Distributional Inclusion Hypothesis and Quantifications}

\label{sec:qtfc_dih}


The Distributional Inclusion Hypothesis (DIH) asserts that the typical characteristic features (contexts) of $r_h$ are expected to appear with $r_H$ if and only if $r_H$ is a hypernym of $r_h$.
Although \citet{geffet-dagan-2005-distributional} report that the DIH is largely valid on a real corpus, it is not deemed fully correct in general as a hyponym can appear in exclusive contexts due to collocational \citep{rimell-2014-distributional} and pragmatic reasons \citep{pannitto2018refining}, and feature inclusion has been found to be selective \citep{roller-etal-2014-inclusive}.
In this section, we describe how quantifications can also be pivotal to the hypothesis.

While \citet{geffet-dagan-2005-distributional} consider syntax-based context, we suggest that contexts based on semantic representation are more suitable since syntactic differences do not necessarily contribute to semantic ones (e.g., passivizations and inversions), and the subject of concern should be semantics.
We use Dependency Minimal Recursion Semantics \citep[DMRS; ][]{copestake2005mrs, copestake-2009-invited} as the semantic representation, which is derived using the English Resource Grammar \citep[ERG; ][]{flickinger2000erg, flickinger2011erg}.
\autoref{fig:pgm} shows the predicate--argument structure of an example DMRS graph.
If $r_i \xleftarrow{\textsc{arg[}a\textsc{]}} r_j$ exists in the DMRS graph of a sentence in the corpus,
we can say that $r_i$ appears in the context $\xleftarrow{\textsc{arg[}a\textsc{]}} r_j$.

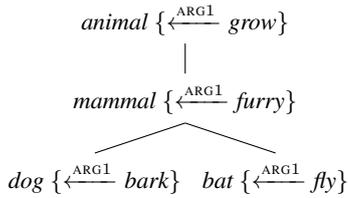
\begin{figure}[t]
    \centering
    \small
    \begin{tikzpicture}
    \Tree [
        .\textit{animal $\{\xleftarrow{\textsc{arg}1}\textit{grow}\}$}
        [.\textit{mammal $\{\xleftarrow{\textsc{arg}1}\textit{furry}\}$}
            \textit{dog $\{\xleftarrow{\textsc{arg}1}\textit{bark}\}$}
            \textit{bat $\{\xleftarrow{\textsc{arg}1}\textit{fly}\}$}
        ]
    ]
    \end{tikzpicture}
    \caption{A taxonomic hierarchy of nouns. Next to each noun is the set of contexts that are applicable to the extension of it and those of its descendants (e.g., all dogs are furry, but not all animals.).}
    \label{fig:hier}
\end{figure}

\begin{table}[t]
  \small
  \centering
  \begin{threeparttable}
    \begin{tabular}{l}
    \toprule  
    \textbf{Corpus 1 (DIH)} \\
    \midrule
    \textit{a dog barks} \\
    \textit{a mammal barks} \\
    \textit{an animal barks} \\
    \textit{a bat flies} \\
    \textit{a mammal flies} \\
    \textit{an animal flies} \\
    \textit{a mammal is furry} \\
    \textit{an animal is furry} \\
    \textit{an animal grows} \\
    \bottomrule  
    \end{tabular}
  \end{threeparttable}
  \qquad
  \begin{threeparttable}
    \begin{tabular}{l}
    \toprule  
    \textbf{Corpus 2 (rDIH)} \\
    \midrule
    \textit{every dog barks} \\
    \textit{every dog is furry} \\
    \textit{every dog grows} \\
    \textit{every bat flies} \\
    \textit{every bat is furry} \\
    \textit{every bat grows} \\
    \textit{every mammal is furry} \\
    \textit{every mammal grows} \\
    \textit{every animal grows} \\
    \bottomrule  
    \end{tabular}
  \end{threeparttable}
  \caption{Corpora generated from the hierarchy in \autoref{fig:hier}. Existential and universal quantifications result in two corpora that follow the DIH and rDIH respectively.} 
  \label{tab:true_snt}
\end{table}

Consider a corpus as a partial description of
a world.
Distributional properties would depend on how the world is described.
Here, we consider a corpus of simple sentences in the form `\textit{[quantifier] [noun] [context word]}'.
Take the taxonomic hierarchy in \autoref{fig:hier} as an example,
where each noun has a set of applicable contexts.
If we want to generate existentially quantified statements that are true,
then: (1) a noun can appear in its hypernyms' contexts,
e.g., `\textit{a dog grows}', where $\xleftarrow{\textsc{arg1}} \textit{grow}$ is applicable to \textit{animal};
and (2) a noun can appear in its hyponyms' contexts,
e.g., `\textit{an animal barks}', where $\xleftarrow{\textsc{arg1}} \textit{bark}$ is applicable to \textit{dog}.
If we only generate (2) and restrict (1) so that contexts that are broadly applicable are not used with more specific nouns, this creates a corpus that follows the DIH.
Corpus~1 of \autoref{tab:true_snt} shows an example.\footnote{Without the restriction on (1), exhaustively generating true assertions generates a corpus where the DIH does not hold between nouns in a unary chain (e.g., \textit{animal} and \textit{mammal} in \autoref{fig:hier}), which would appear in the same set of contexts.}

In contrast, generating sentences with universal quantifications results in a corpus that follows the \textit{reverse} of the DIH (rDIH), as in Corpus~2, where the set of contexts of \textit{mammal} is a subset of that of \textit{dog}.
Consequently, methods that rely on the DIH as a cue for hypernymy would be undermined.

In \autoref{sec:exp_real}, we use these processes to generate
corpora which strictly align with the DIH or rDIH.
Corpora with more complex sentence structures
would require a richer world model than can be encoded in
a taxonomic hierarchy like \autoref{fig:hier}.
For instance, with a restricted relative clause, `\textit{every dog that is trained is gentle}' does not entail `\textit{every Chihuahua is gentle}' even if \textit{Chihuahua} is a hyponym of \textit{dog}, as the universal quantifier applies only to trained dogs.
We also disregard negations because they can co-occur nearly freely, effectively making a context word in the negated scope uninformative.
For example, `\textit{a dog does not \rule{0.7cm}{0.1mm}}' is much less selective than `\textit{a dog \rule{0.7cm}{0.1mm}}'.

\section{Functional Distributional Semantics}
\label{sec:fds}

In this section, we introduce Functional Distributional Semantics (FDS), discuss hypernymy representation in FDS and explain how FDS can be adapted to handle quantifications.
We follow \citet{lo-etal-2023-functional}'s FDS implementation which is briefly described here.

\subsection{Model-Theoretic Semantics}
\label{subsec:mts}
FDS is motivated by model-theoretic semantics, which sees meaning in terms of an extensional model structure that consists of a set of \textit{entities}, and a set of \textit{predicates}, each of which is true or false of the entities.
In parallel, FDS represents an entity by a \textit{pixie} which is taken to be a high-dimensional feature vector, and represents a predicate by a truth-conditional \textit{semantic function} which takes pixie(s) as input and returns the probability of truth.

\subsection{Probabilistic Graphical Models}
\label{subsec:pgm}

\begin{figure}[t]
\centering
\includegraphics[width=0.9\linewidth]{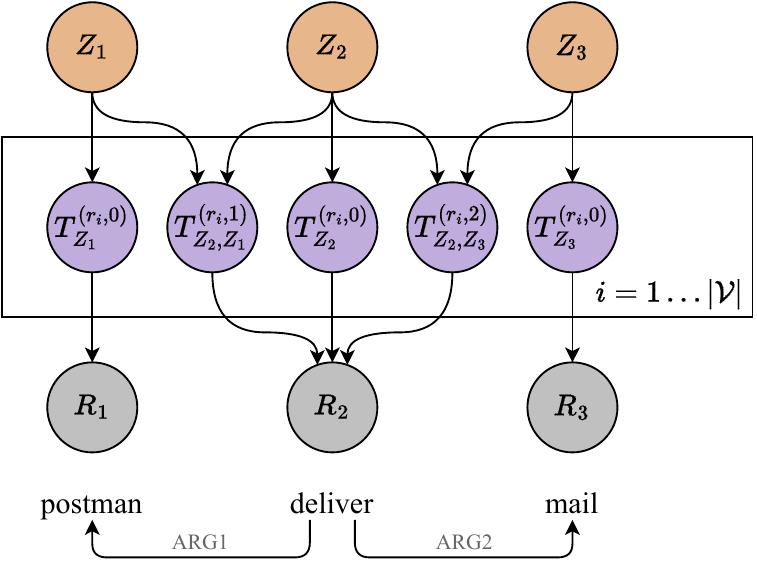}
\caption{Probabilistic graphical model of FDS for generating words in an SVO triple ‘\textit{postman deliver mail}'.
Only $R_1 = \textit{postman}$, $R_2 = \textit{deliver}$, and $R_3 = \textit{mail}$ are observed.
}
\label{fig:pgm}
\end{figure}

The framework is formalized in terms of a family of probabilistic graphical models.
Each of them describes the generative process of predicates in the semantic graph of a sentence.
\autoref{fig:pgm} illustrates the process of generating the words given the argument structure $R_1 \xleftarrow{\textsc{arg}1} R_2 \xrightarrow{\textsc{arg}2} R_3$.
First, a pixie $Z_j \in \mathbb{R}^d$ is generated for each node in the graph, together representing the entities described by the sentence.
Then, for each pixie $Z_j$, a truth value $T^{(r_i, 0)}_{Z_j}$ is generated for each predicate $r_i$ in the vocabulary $\mathcal{V}$; and for each pair of nodes connected as $R_j \xrightarrow{\textsc{arg}a} R_k$ whose corresponding pixies are $Z_j$ and $Z_k$, a truth value $T^{(r_i, a)}_{Z_j, Z_k}$ is generated for each predicate $r_i$ in the vocabulary.
Finally, a single predicate $R_j$ is generated for each pixie $Z_j$ conditioned on the truth values.

\subsection{Semantic Functions}
\label{subsec:sem_func}
As mentioned in \autoref{subsec:mts}, instead of treating a predicate as an indicator function, FDS models the probability that it is true of the pixie(s) with unary (in \eqref{eq:unary_semfunc}) and binary semantic functions (in \eqref{eq:binary_semfunc}). This allows the model to
account for vagueness.
\begin{align}
\label{eq:unary_semfunc}
P\left(T^{(r_i,0)}_{Z_j} {=} \top \,\middle|\, z_j\right) &= t^{(r_i,0)}(z_j) \\
\label{eq:binary_semfunc}
P\left(T^{(r_i,a)}_{Z_j, Z_k} {=} \top \,\middle|\, z_j, z_k\right) &= t^{(r_i,a)}(z_j, z_k)
\end{align}
The functions are implemented as linear classifiers:
\begin{gather}
\label{eq:sem_func1}
t^{(r_i, 0)}(z_j) = S \left( {v^{(r_i, 0)}}^\top z_j + b^{(r_i, 0)} \right) \\
\label{eq:sem_func2}
\begin{multlined}
t^{(r_i, a)}(z_j, z_k) = \\
S \left( {v^{(r_i, a)}_1}^\top z_{j} + {v^{(r_i, a)}_2}^\top z_{k} + b^{(r_i, a)} \right)
\end{multlined}
\end{gather}
where $S$ denotes the sigmoid function.





\subsection{Representing Hypernymy}
\label{subsec:rep_hyp}


In truth-conditional semantics, for a set of entities $D$, $r_H$ is a hypernym of $r_h$ if and only if
\begin{equation}
    \label{eq:hyp_def}
    \forall x {\in} D {\colon} r_h(x) {\implies} r_H(x)
\end{equation}
Although FDS provides truth-conditional interpretations of words, it is not straightforward to define hypernymy in FDS where predicates are probabilistic and work over high-dimensional pixies.
One way is to translate \eqref{eq:hyp_def} to probabilistic counterpart for a score on hypernymy $P\left(T^{(r_H,0)}_{Z} {=} \top \,\middle|\, T^{(r_h,0)}_{Z} {=} \top\right)$.
However, this conditional probability is unavailable since only $P\left(T^{(r_H,0)}_{Z} {=} \top  \,\middle|\, z \right)$ and $P\left(T^{(r_h,0)}_{Z} {=} \top \,\middle|\, z \right)$ are modelled by FDS.

Another way is to interpret the probability model from a fuzzy set perspective and use fuzzy set containment \citep{zadeh1965fuzzy} for representing hypernymy:
\begin{equation}
    \label{eq:fds_hyp_def}
    \forall z \colon t^{(r_H, 0)}(z) \;{>}\; t^{(r_h, 0)}(z)
\end{equation}
Note that if we consider all $z \in \mathbb{R}^d$, \eqref{eq:fds_hyp_def} can only be true when $v^{(r_h, 0)} = kv^{(r_H, 0)}$ where $k \neq 0$, which is impossible to be obtained in practice from model training.
Therefore, we restrict the pixie space and only consider pixies in a unit hypersphere or hypercube to be meaningful. With \eqref{eq:sem_func1} and \eqref{eq:sem_func2}, $r_H$ is considered the hypernym of $r_h$ if and only if $s(r_h, r_H) > 0$ in \eqref{eq:fds_hyp_def_l1l2}, where $p \in \{1,2\}$ (derivation in \autoref{app:derivation_hyp_def}).
\citet{cheng-etal-2023-embedded} also use this score for hypernymy.
\begin{equation}
    \begin{aligned}
    \label{eq:fds_hyp_def_l1l2}
    s(r_h, r_H) = \; & b^{(r_H, 0)} - b^{(r_h, 0)} \\
                   & - {\left \lVert v^{(r_H, 0)} - v^{(r_h, 0)} \right \rVert}_p
    \end{aligned}
\end{equation}
Note that the transitivity of \eqref{eq:hyp_def} is paralleled (derivation in \autoref{app:derivation_trans}):
\begin{equation}
    \label{eq:transitve}
    \begin{multlined}
    s(r_1, r_2) > 0 \land s(r_2, r_3) > 0 \\
    \implies s(r_1, r_3) > 0
    \end{multlined}
\end{equation}

Having hypernymy representation built into a distributional model allows generalization out of missing information.
For example, when the (r)DIH does not hold between \textit{dog} and \textit{mammal} in a corpus, knowing that both \textit{dog} and \textit{fox} share the same contexts in a corpus (e.g., $\{\xleftarrow{\textsc{arg}1}\textit{bark}\}$) is indicative that they share common hypernyms, e.g., \textit{mammal}.
This also applies to hyponymy, e.g., \textit{machine} and \textit{system} sharing $\{\xleftarrow{\textsc{arg}1}\textit{complex}\}$ and sharing \textit{computer} as their hyponym.
Such generalization power is largely absent in models based on strict contexts subsumption.




\subsection{Original Training Objective}
\label{subsec:model_train}

FDS models are trained using the variational-autoencoding method on simplified DMRS graphs where quantifiers and scopal information are removed from the graphs before training, leaving us with just the predicate--argument structure.
The approximate posterior distribution of pixies $q_\phi$ is taken to be $n$ spherical Gaussian distributions, each with mean $\mu_{Z_i}$ and covariance $\sigma_{Z_i}^2I$.
Given an observed DMRS graph $G$ with $n$ pixies $Z_1 \dots Z_n$, we maximize \eqref{eq:final_loss}, reformulated from the $\beta$-VAE \citep{higgins2017betavae}.
\begin{align}
    \phantom{\mathcal{C}_{i,j,a}}
    & \begin{aligned}
    \mathllap{\mathcal{L}} = & \sum_{i=1}^n \mathcal{C}_i + \sum_{r_i \xrightarrow{\textsc{arg}[a]} r_j \textup{in } G} \mathcal{C}_{i,j,a} \\
    & - \frac{d}{2} \sum_{i=1}^n \beta_1 \mu_{Z_i}^2 + \beta_2 \left ( \sigma_{Z_i}^2 - \ln{\sigma_{Z_i}^2} \right )
    \end{aligned} \label{eq:final_loss}
\end{align}
The last term in \eqref{eq:final_loss} is the regularization term for the approximate posterior, and
the first two terms aim to maximize the truthness of observed predicates and the falsehood of $K$ negatively sampled ones $r_k'$ over the inferred pixie distribution $q_\phi$, by
\begin{align}
    \phantom{\mathcal{C}_{i,j,a}}
    & \begin{multlined}
        \mathllap{\mathcal{C}_i} = \ln\mathbb{E}_{q_\phi}\left[t^{(r_i,0)}(z_i)\right] \\
        +\sum_{k=1}^K \ln\mathbb{E}_{q_\phi}\left[1-t^{(r_k',0)}(z_i)\right]
    \end{multlined} \label{eq:neg_samp_arg0} \\
    & \begin{multlined}
        \mathllap{\mathcal{C}_{i,j,a}} = \ln\mathbb{E}_{q_\phi}\left[t^{(r_i,a)}(z_i, z_j)\right] \\
        +\sum_{k=1}^K \ln\mathbb{E}_{q_\phi}\left[1-t^{(r_k',a)}(z_i, z_j)\right]
    \end{multlined} \label{eq:neg_samp_arga}
\end{align}

Both the local predicate–argument structure of each predicate and global topical information in the graph are used for variational inference.
For instance, the approximate posterior distribution of the pixie $Z_1$ of \textit{postman} in \autoref{fig:pgm} is inferred from the direct argument information, $\xleftarrow{\textsc{arg}1} \textit{deliver}$, and the indirect topical predicate, $\not \xleftarrow{} \textit{mail}$.
\paragraph{Our Hypothesis.}
If the training corpus strictly follows the DIH, hypernymy can be learnt by FDS models.
The intuition behind our hypothesis is elaborated in \autoref{app:intuition}.

\subsection{Proposed Objective for Universal Quantifications}
\label{subsec:uni_obj}
FDS assumes that each observed predicate refers to only one point in the pixie space and offers no tools for dealing with regions.
We propose a method to allow optimizations of semantic functions with respect to a region in the pixie space, thus enabling FDS to handle simple sentences with universal quantifications.
Essentially, we add the following \bm{$\forall$}\textbf{-objective} to the original objective in \eqref{eq:final_loss}:
\begin{equation}
    \begin{aligned}
        \label{eq:uni_obj}
        \mathcal{L}_\forall = \sum_{ r_j \xleftarrow{\textsc{arg}[a]} r_i ~\textup{in}~G} {s_a(r_i, r_j) + \mathcal{U}_{ i,j,a }}
    \end{aligned}
\end{equation}
where $r_j$ is a predicate whose referent is universally quantified, and
\begin{align}
    & \begin{aligned}
        s_a(r_i, r_j) = \; & b^{(r_i, a)} - b^{(r_j, 0)} \\
               & - {\left \lVert v^{(r_i, a)}_2 - v^{(r_j, 0)} \right \rVert}_p
    \end{aligned} \label{eq:uni_obj_pos} \\
    &  \begin{multlined}
        \mathcal{U}_{ i,j,a } = \sum_{k=1}^K  \min \left (0, -s_0(r_i, r_k') \right ) \\
        + \sum_{k=1}^K \min \left (0, -s_{a_k''}(r_k'', r_j) \right )
    \end{multlined} \label{eq:uni_obj_neg}
\end{align}
Note that \eqref{eq:uni_obj_pos} is modified based on \eqref{eq:fds_hyp_def_l1l2}, previously defined for classifying hypernymy.

To explain \eqref{eq:uni_obj}, consider the sentence `\textit{every dog barks}' as an example.
The first term inside the summation in \eqref{eq:uni_obj} enforces that extension of $r_j$ is a subset of that of prototypical argument $a$ of $r_i$, i.e., the set of dogs should be contained in the set of agents that barks.
The second term, described in \eqref{eq:uni_obj_neg}, incorporates $K$ randomly generated negative samples.
$r_k'$ is a noun, which is a negative sample for $r_j$. $r_k''$ is a verb or adjective and $a_k''$ is an argument role, together form a negative sample for $r_i$ and $a$.
Then, \eqref{eq:uni_obj_neg} requires that it is false to universally quantify the referents of the noun $r_k'$ in $r_k' \xleftarrow{\textsc{arg}[a]} r_i$ and $r_j$ in $r_j \xleftarrow{\textsc{arg}[a_k'']} r_k''$.
For the example, both of the following sentences are considered false:
`\textit{every dog is owned}' and `\textit{every cat barks}', where $r_k' = \textit{cat}$, $ r_k''=\textit{own}$ and $a_k'' = 2$.




\section{Experiments on Synthetic Data Sets}
\label{sec:exp_synth}

Testing our hypothesis in \autoref{subsec:model_train} and the effectiveness of the new objective for universal quantifications in \autoref{subsec:uni_obj} requires corpora that strictly follow the DIH or rDIH, which is impractical for real corpora.
Therefore, we create a collection of synthetic data sets and perform experiments on them.



\begin{figure*}[t]

    \centering
    \begin{subfigure}[t]{0.24\linewidth}
        \centering
        \begin{tikzpicture}[level distance=25, scale = 0.8]
        \scriptsize
        \small
        \tikzset{
            every tree node/.style={anchor=north, align=center},
        }
        \Tree [
            .{$r_1$ $\{c_1\}$}
            [
                .{$r_2$ $\{c_2\}$}
                {$r_3$ $\{c_3\}$}
            ]
        ]
        \begin{scope}[xshift=1.1cm]
        \Tree [
            .{$r_4$ $\{c_4\}$}
            [
                .{$r_5$ $\{c_5\}$}
                {$r_6$ $\{c_6\}$}
            ]
        ]
        \end{scope}
        \begin{scope}[xshift=2.2cm]
        \Tree [
            .{$r_7$ $\{c_7\}$}
            [
                .{$r_8$ $\{c_8\}$}
                {$r_9$ $\{c_9\}$}
            ]
        ]
        \end{scope}
        \begin{scope}[xshift=3.4cm]
        \Tree [
            .{$r_{10}$ $\{c_{10}\}$}
            [
                .{$r_{11}$ $\{c_{11}\}$}
                {$r_{12}$ $\{c_{12}\}$}
            ]
        ]
        \end{scope}
        \end{tikzpicture}
        \caption{Four chains ($H_\textup{chains}$)}
        \label{subfig:g_chains}
    \end{subfigure}
    \hfill
    \begin{subfigure}[t]{0.18\linewidth}
        \centering
        \begin{tikzpicture}[level distance=25, scale = 0.8]
            \small
            \tikzset{
                every tree node/.style={anchor=north, align=center},
                level 1/.style={sibling distance=-2pt},
                level 2/.style={sibling distance=-2pt},
            }
            \Tree [
                .{$r_1$ $\{c_1\}$}
                [
                    .{$r_2$ $\{c_2\}$}
                    {$r_4$ $\{c_4\}$}
                    \node (A) {$r_5$ $\{c_5\}$};
                ]
                [
                    .\node (B) {$r_3$ $\{c_3\}$};
                    {$r_6$ $\{c_6\}$}
                ]
            ]
        \end{tikzpicture}
        \caption{Tree ($H_\textup{tree}$)}
        \label{subfig:g_tree}
    \end{subfigure}
    \hfill
    \begin{subfigure}[t]{0.18\linewidth}
        \centering
        \begin{tikzpicture}[level distance=25, scale = 0.8]
            \small
            \tikzset{
                every tree node/.style={anchor=north, align=center},
                level 1/.style={sibling distance=-2pt},
                level 2/.style={sibling distance=-2pt},
            }
            \Tree [
                .{$r_1$ $\{c_1\}$}
                [
                    .{$r_2$ $\{\underline{c_2}\}$}
                    {$r_4$ $\{c_4\}$}
                    \node (A) {$r_5$ $\{c_5\}$};
                ]
                [
                    .\node (B) {$r_3$ $\{c_3\}$};
                    {$r_6$ $\{\underline{c_2}\}$}
                ]
            ]
        \end{tikzpicture}
        \caption{Tree with overlapping contexts (underlined) ($H_\textup{tree}'$)}
        \label{subfig:g_tree'}
    \end{subfigure}
    \hfill
    \begin{subfigure}[t]{0.18\linewidth}
        \centering
        \begin{tikzpicture}[level distance=25, scale = 0.8]
            \small
            \tikzset{
                every tree node/.style={anchor=north, align=center},
                level 1/.style={sibling distance=-2pt},
                level 2/.style={sibling distance=-2pt},
            }
            \Tree [
                .{$r_1$ $\{c_1\}$}
                [
                    .{$r_2$ $\{c_2\}$}
                    {$r_4$ $\{c_4\}$}
                    \node (A) {$r_5$ $\{c_5\}$};
                ]
                [
                    .\node (B) {$r_3$ $\{c_3\}$};
                    {$r_6$ $\{c_6\}$}
                ]
            ]
            \draw (A.north) -- (B.south);
        \end{tikzpicture}
        \caption{DAG ($H_\textup{DAG}$)}
        \label{subfig:g_dag}
    \end{subfigure}
    \hfill
    \begin{subfigure}[t]{0.18\linewidth}
        \centering
        \begin{tikzpicture}[level distance=25, scale = 0.8]
            \small
            \tikzset{
                every tree node/.style={anchor=north, align=center},
                level 1/.style={sibling distance=-2pt},
                level 2/.style={sibling distance=-2pt},
            }
            \Tree [
                .{$r_1$ $\{c_1\}$}
                [
                    .{$r_2$ $\{\underline{c_2}\}$}
                    {$r_4$ $\{c_4\}$}
                    \node (A) {$r_5$ $\{c_5\}$};
                ]
                [
                    .\node (B) {$r_3$ $\{c_3\}$};
                    {$r_6$ $\{\underline{c_2}\}$}
                ]
            ]
            \draw (A.north) -- (B.south);
        \end{tikzpicture}
        \caption{DAG with overlapping contexts (underlined) ($H_\textup{DAG}'$)}
        \label{subfig:g_dag'}
    \end{subfigure}
    \caption{Examples of the topologies of the synthetic taxonomic hierarchies.} 
    \label{fig:synth_hier}
\end{figure*}
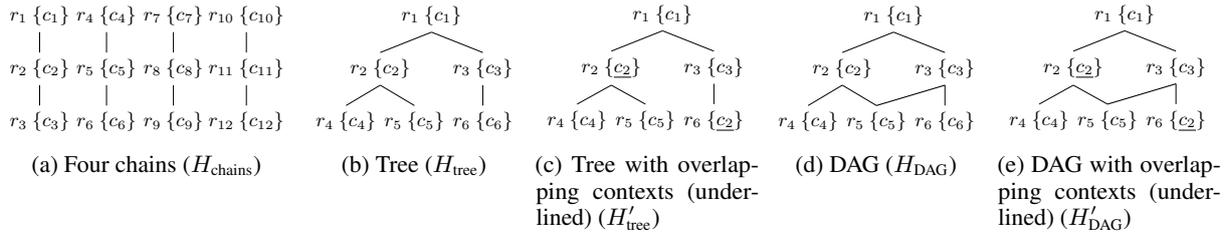

\subsection{Synthetic Data Sets under the (r)DIH}
Each of the synthetic data sets consists of a taxonomic hierarchy of nouns and a corpus, created using the following procedure:
\begin{enumerate}[itemsep=0pt]
  \item \textbf{Create a taxonomic hierarchy.} Define a set of nouns, the hypernymy relations of them, and the contexts applicable to its extension and those of its hyponyms (as in \autoref{fig:hier}).
  \item \textbf{Choose a hypothesis.} The DIH or rDIH.
  \item \textbf{Create a corpus.} Create sentences in the form `\textit{[quantifier] [noun] [context word]}' following the chosen hypothesis and the defined hierarchy (as in \autoref{tab:true_snt}).
\end{enumerate}


\subsubsection{Topology of Hierarchy}
Different topologies of hierarchy lead to different distributional usage of words, thus possibly varying representations learnt for hypernymy.
For example, a noun can have multiple hypernyms (e.g., \textit{dog} is the hyponym of both \textit{pet} and \textit{mammal}), or share overlapping contexts with another noun far in the hierarchy (e.g., both \textit{bat} and \textit{airplane} $\xleftarrow{\textsc{arg}1}\textit{fly}$).

To test the robustness of FDS models for learning hypernymy, we experiment with a range of topologies.
\autoref{fig:synth_hier} exemplifies the five classes of topologies used.
We expect that directed acyclic graphs ($H_{\textup{DAG}}$ and $H_{\textup{DAG}}'$) be harder topologies than trees ($H_{\textup{tree}}$ and $H_{\textup{tree}}'$), and topologies with overlapping contexts ($H_{\textup{tree}}'$ and $H_{\textup{DAG}}'$) be harder than those without ($H_{\textup{tree}}$ and $H_{\textup{DAG}}$).
In addition, we test $H_\textup{chains}$ with pixie dimensionality $d=2$.
A 2-D pixie space
allows lossless visualization of the semantic functions.
To test hypernymy learning at scale on an actual hierarchy, we make use of WordNet \citep{miller1995wordnet, fellbaum1998wordnet} and test our models on the WordNet's hierarchy ($H_{\textup{WN}}$).


Every node in the hierarchy consists of a noun and a semantic context.
The topology of the $H_{\textup{chains}}$ used in the experiment is exactly as depicted in \autoref{fig:synth_hier}.
$H_{\textup{WN}}$ is created out of the synset \textit{entity.n.01} in WordNet, which is the root, and its hyponymic synsets. This results in 74,374 nodes with 663,492 hypernymy pairs.
We randomly sample 663,492 pairs from the remaining pairs as negative instances for evaluation.
For the remaining hierarchies, each of them consists of 153 nodes with a height of 5.
For $H_{\textup{tree}}$, the first level is a root node, and a node at the $h$\textsuperscript{th} level has $(h+1)$ direct children.
$H_{\textup{tree}'}$ is created from $H_{\textup{tree}}$ by choosing 5 pairs of nodes and making each pair share a context set.
$H_{\textup{DAG}}$ and $H_{\textup{DAG}}'$ are created from $H_{\textup{tree}}$ and $H_{\textup{tree}'}$ respectively by choosing 5 pairs of nodes, where the nodes of each pair are at different levels, and make the higher level node the direct parent of the lower level one. 

\subsection{FDS Models Training}
\label{subsec:model_train_synth}
We experiment with two variations of FDS training: \textsc{Fds} is trained using the original objective in \eqref{eq:final_loss} whereas \textsc{Fds}\textsubscript{$\forall$} incorporates the $\forall$-objective following \autoref{subsec:uni_obj}.
The hypernymy score of each model, given by \eqref{eq:fds_hyp_def_l1l2}, is averaged over two runs of different random seed.
We empirically find that setting $p = 1$ in \eqref{eq:uni_obj_pos} and $p = 2$ in \eqref{eq:fds_hyp_def_l1l2} almost always give the best performances, and we only report the results in this setup.
Other than the newly introduced training objective, training of the models largely follows that of \citet{lo-etal-2023-functional}.
No hyperparameter search is conducted due to the large number of experiments (details described in \autoref{app:train}).

\subsection{Evaluation on Hypernymy Detection}
\label{subsec:synth_eval1}

We test if a model trained on the corpus learns to identify hypernymy defined in the hierarchy that generates the corpus.
Concretely, a model is asked to give a score of hypernymy between every pair of nouns using \eqref{eq:fds_hyp_def_l1l2}. Performance is then measured by the area under the receiver operating characteristic curves (AUC).
Unlike average precision, AUC values do not reflect changes in the distribution of classes, which is favourable since we are comparing models' performances across varying class distributions generated from different topologies.

We include two distributional methods for hypernymy detection based on the DIH in the experiments, namely WeedsPrec \citep{weeds-etal-2004-characterising} and invCL \citep{lenci-benotto-2012-identifying}:
\begin{equation}
    \label{eq:weedsprec}
    \textup{WeedsPrec}(r_1, r_2) = \frac{ \sum_i u^{(r_1)}_i \mathbbm{1}_{u^{(r_2)}_i > 0}}{ \sum_i u^{(r_1)}_i} \nonumber
\end{equation}
\begin{equation}
    \label{eq:invcl}
    \begin{aligned}
        & \textup{invCL}(r_1, r_2) = \sqrt{
            \textup{CL}(r_1, r_2)(1 - \textup{CL}(r_2, r_1))
        } \\
        & \textup{where}~\textup{CL}(r_1, r_2) = \frac{
            \sum_i \min \left ( u^{(r_1)}_i, u^{(r_2)}_i \right )}{ \sum_i u^{(r_1)}_i} 
        \nonumber
    \end{aligned}
\end{equation}
They measure the context inclusion of $r_1$ by $r_2$ and invCL measures also the non-inclusion of $r_2$ by $r_1$.
Their distributional space is constructed by first counting co-occurrences of adjacent predicates in the preprocessed DMRS graphs, then transforming the resulting matrix using positive pointwise mutual information.
Each row vector $u^{(r_i)}$ in the transformed matrix represents a predicate $r_i$.

\begin{table}[th]
  \small
  \centering
  \setlength{\tabcolsep}{0.1cm}
  \begin{threeparttable}
    \begin{tabular}{lcccccc}
    \toprule  
    \textbf{Model} & $H_\textup{chains}$ & $H_\textup{tree}$ & $H_\textup{tree}'$ & $H_\textup{DAG}$ & $H_\textup{DAG}'$ & $H_\textup{WN}$ \\
    \midrule
\textsc{Fds} & .990 & .994 & .995 & .995 & .995 & .940 \\
\textsc{Fds}\textsubscript{$\forall$} & .925 & .206 & .210 & .214 & .221 & .788 \\
WeedsPrec & 1.000 & 1.000 & 1.000 & 1.000 & 1.000 & 1.000 \\
invCL & 1.000 & 1.000 & 1.000 & 1.000 & .999 & 1.000 \\
    \bottomrule  
    \end{tabular}
  \end{threeparttable}
  \caption{AUC of models trained on the DIH corpora.} 
  \label{tab:synth_dih}
\end{table}

\begin{table}[th]
  \small
  \centering
  \setlength{\tabcolsep}{0.1cm}
  \begin{threeparttable}
    \begin{tabular}{lcccccc}
    \toprule  
    \textbf{Model} & $H_\textup{chains}$ & $H_\textup{tree}$ & $H_\textup{tree}'$ & $H_\textup{DAG}$ & $H_\textup{DAG}'$ & $H_\textup{WN}$ \\
    \midrule
\textsc{Fds} & .876 & .842 & .793 & .752 & .688 & .444 \\
\textsc{Fds}\textsubscript{$\forall$} & .988 & .983 & .978 & .981 & .977 & .675 \\
WeedsPrec & .900 & .675 & .619 & .613 & .556 & .809 \\
invCL & .900 & .355 & .280 & .236 & .276 & .564 \\
    \bottomrule  
    \end{tabular}
  \end{threeparttable}
  \caption{AUC of models trained on the rDIH corpora.} 
  \label{tab:synth_rdih}
\end{table}

\autoref{tab:synth_dih} and \autoref{tab:synth_rdih} show the results of FDS models when trained on the DIH and rDIH corpora respectively (a visualization of results on $H_\textup{chains}$ is provided in \autoref{app:vis}).
As expected, \textsc{Fds}, WeedsPrec and invCL are shown to work on the DIH corpora, and only \textsc{Fds}\textsubscript{$\forall$} works on the rDIH corpora.
Reversing the FDS models on respective corpora yields substantially worse performances.
In particular, \textsc{Fds}\textsubscript{$\forall$} attains AUCs of about 0.2 on the DIH corpora means hypernymy predictions are even mostly reversed,
which in turn reflects the effectiveness of the universal objective when \textsc{Fds}\textsubscript{$\forall$} interprets the subsumption of contexts reversely based on the rDIH.
Moreover, hierarchies with overlapping contexts and multiple direct hypernyms are not harder for FDS than those without.
Scaling up to the huge WordNet hierarchy $H_\textup{WN}$ results in a slight drop in AUC for \textsc{Fds} on the DIH corpus, and markedly worse performances for \textsc{Fds}\textsubscript{$\forall$}.
While our preset hyperparameters work nicely on all other settings, it is possible that \textsc{Fds}\textsubscript{$\forall$} requires a different set of hyperparameters to perform optimally on the rDIH corpora generated from huge hierarchies.

\begin{table}[th]
  \small
  \centering
  \begin{threeparttable}
    \begin{tabular}{lcccc}
    \toprule  
    \textbf{Model} & $H_\textup{chains}$ & $H_\textup{tree}$ & $H_\textup{DAG}$ & $H_\textup{WN}$ \\
    \midrule
    \textsc{Fds} & .947 & .991 & .990 & .946 \\
    \textsc{Fds}\textsubscript{$\forall$} & .956 & .358 & .378 & .183 \\
    \textsc{Fds}\textsubscript{($\forall$)} & .999 & .986 & .984 & .992 \\
    WeedsPrec & .950 & .996 & .993 & .993 \\
    invCL & .958 & .891 & .864 & .868 \\
    \bottomrule  
    \end{tabular}
  \end{threeparttable}
  \caption{AUC of models trained on combined corpora.} 
  \label{tab:synth_mixed}
\end{table}

We further produce new corpora by combining the DIH and rDIH corpus of each topology.
The resulting corpora still follow the DIH.
In this setup, instead of applying the same FDS training objective across the whole corpus, the $\forall$-objective can be added only when there is a universal quantifier (e.g., \textit{every}).
We name this FDS model \textsc{Fds}\textsubscript{($\forall$)}.
\autoref{tab:synth_mixed} shows the results.
The DIH methods (\textsc{Fds}, WeedsPrec, and invCL) perform well as expected, whereas invCL performs worse since it also measures non-inclusion which is undermined by the rDIH half.
There are several interesting insights into FDS models.
First, \textsc{Fds} still works on the corpora with an rDIH half because it is still valid to say `\textit{a dog grows}', as mentioned in \autoref{sec:qtfc_dih}.
Second, \textsc{Fds}\textsubscript{($\forall$)} is as good as \textsc{Fds} across topologies and even better on $H_\textup{chains}$ and $H_\textup{WN}$.
This implies that the $\forall$-objective indeed captures simple universal quantifications and can be used compatibly with the original training method on a corpus with varying quantifications.
Third, on $H_\textup{WN}$, \textsc{Fds}\textsubscript{($\forall$)} performs better than \textsc{Fds} on the DIH corpora and much better than \textsc{Fds}\textsubscript{$\forall$} on the rDIH corpora.
This reflects that the $\forall$-objective is more effective when mixed with the original mode of training.

\subsection{Evaluation on Distributional Generalization}
\label{subsec:synth_eval2}

\begin{figure}[t]
    \centering
    \subcaptionbox{An example taxonomy with missing relations. The siblings $r_4$ and $r_5$ have the same context $c_5$. Dashed lines show the relations to be removed from training.\label{subfig:dist_gen}}[0.46\linewidth]{
        \centering
        \begin{tikzpicture}[level distance=25, scale = 0.8]
            \small
            \tikzset{
                every tree node/.style={anchor=north, align=center},
                level 1/.style={sibling distance=-2pt},
                level 2/.style={sibling distance=-2pt},
            }
            \Tree [
                .{$r_1$ $\{c_1\}$}
                [
                    .\node (B) {$r_2$ $\{c_2\}$};
                    \edge[dashed] {};
                    [
                        .\node (C) {$r_4$ $\{\underline{c_5}\}$};
                        \node (D) {$r_7$ $\{c_7\}$};
                    ]
                    [
                        .\node (E) {$r_5$ $\{\underline{c_5}\}$};
                        \node (G) {$r_8$ $\{c_8\}$};
                    ]
                ]
                [
                    .\node (F) {$r_3$ $\{c_3\}$};
                    \node (A) {$r_6$ $\{c_6\}$};
                ]
            ]
            \draw (A.north) -- (B.south);
            \draw[dashed] (D.north) -- (E.south);
            \draw (C.north) -- (F.south);
        \end{tikzpicture}
    }
    \hfill%
    \subcaptionbox{Contexts that $r_4$ and $r_5$ in \subref{subfig:dist_gen} would appear with in the new (r)DIH corpora.\label{subtab:dist_gen}}[0.46\linewidth]{
        \centering
        \small
        \setlength{\tabcolsep}{0.1cm}
        \begin{tabular}{lcc}
            \toprule  
            \textbf{Hyp.} & \textbf{Noun} & \textbf{Contexts} \\
            \midrule
            \multirow{2}{*}{DIH} &  
            $r_4$ & $\{c_5, c_7\}$ \\
            & $r_5$ & $\{c_5, c_8\}$ \\
            \midrule
            \multirow{2}{*}{rDIH} &  
            $r_4$ & $\{c_1, c_3, c_5\}$ \\
            & $r_5$ & $\{c_1, c_2, c_5\}$ \\
            \bottomrule  
        \end{tabular}
    }
    \caption{Illustration of the setup for testing distributional generalization.}
    \label{fig:dist_gen}
\end{figure}

We also test if the distributional generalization power mentioned in \autoref{subsec:rep_hyp} exists in FDS.
We construct a new corpus from a hierarchy with removed hypernymy information.
\autoref{fig:dist_gen} illustrates the idea with an example hierarchy of nouns and the contexts that would appear with the nouns in the new corpora obtained.
If upward (downward) distributional generalization exists in a model, based on that $r_4$ and $r_5$ share $c_5$ as their contexts, it should identify the hypernyms (hyponyms) of $r_5$ ($r_4$) as the \textit{candidate hypernyms (hyponyms)} of $r_4$ ($r_5$) after training on the new corpus.
That is, we expect $\forall r_j \in \{r_5,r_6,r_7,r_8\} \colon s(r_4, r_2) > s(r_4, r_j)$ if upward generalization exists, and $\forall r_j \in \{r_1,r_2,r_3,r_4,r_6\} \colon s(r_7, r_5) > s(r_j, r_5)$ if downward exists in FDS.

In our experiments, we sample five nouns from the $H_\textup{DAG}'$ hierarchy.
Then, for each of these nouns $\tilde{r}$, we equate the contexts set of $\tilde{r}$ to that of one of its siblings and remove the hypernymy (hyponymy) information of their common parent (daughter) from $\tilde{r}$ when creating the new corpora.


\begin{table}[h]
  \small
  \centering
  \begin{threeparttable}
    \begin{tabular}{llcc}
    \toprule  
    \textbf{Model} & \textbf{Hypothesis} & Upward & Downward \\
    \midrule
    \textsc{Fds} & DIH & .922 & .742 \\
    \textsc{Fds}\textsubscript{$\forall$} & rDIH & .976 & .998 \\
    \bottomrule  
    \end{tabular}
  \end{threeparttable}
  \caption{Mean AUC for distributional generalizations.}
  \label{tab:synth_eval2}
\end{table}

For each $\tilde{r}$, we compute the hypernymy score of between $\tilde{r}$ and each of the candidate hypernyms, and between $\tilde{r}$ and a random noun.
We measure the performance with mean AUC, averaged over the five chosen $\tilde{r}$.
\autoref{tab:synth_eval2} shows that both upward and downward distributional generalizations exist when the corpus follows either the DIH or rDIH, and to a larger extent on the rDIH corpus.

\subsection{Summary}
The experimental results confirm that: (1) the original FDS models learn hypernymy under the DIH, (2) the proposed $\forall$-objective captures universal quantifications and enables hypernymy learning under the rDIH, and (3) FDS models can generalize about nouns with incomplete contexts in a corpus using distributional information.





\section{Experiments on Real Data Sets}
\label{sec:exp_real}


Seeing how FDS performs on restricted synthetic data sets is helpful for understanding models' behaviour but it does not immediately tell us more about hypernymy learning from open classes of sentences. Therefore, we perform further experiments using a real corpus and data sets for hypernymy.

\subsection{FDS Models Training}
\label{subsec:train_data}

\paragraph{Training Data.}
FDS models are trained on Wikiwoods
\citep{flickinger-etal-2010-wikiwoods, solberg2012corpus}, which provide linguistic analyses of 55m sentences (900m tokens) in English Wikipedia. Each of the sentences was parsed by the PET parser \citep{callmeier2001efficient, toutanova2005stochastic} using the 1212 version of the ERG, and the parses are ranked by a ranking model trained on WeScience \citep{ytrestol2009extracting}.
We extract the DMRS graphs from Wikiwoods using Pydelphin\footnote{\url{https://github.com/delph-in/pydelphin}} \citep{copestake-etal-2016-resources}.
After preprocessing, there are 36m sentences with 254m tokens.

\paragraph{Model Configurations.}
Although quantifications are annotated in Wikiwoods, neither of the proposed training objectives is entirely applicable in general.
For example, even for a sentence of modest complexity like `\textit{every excited dog barks}', it requires a universal quantification over the intersection of the set of dogs and excited entities.
However, set intersection is not modelled by FDS.
In our experiments, we apply either \textsc{Fds} or \textsc{Fds}\textsubscript{$\forall$} described in \autoref{subsec:model_train_synth} to every training instance.
We also test an additional model \textsc{Fds}\textsubscript{$\forall$/2} where the $\forall$-objective is scaled by 0.5.
Each model is trained for 1 epoch and the results of each model are averaged over two random seeds as discussed in \autoref{subsec:model_train_synth}.

\subsection{Evaluation Method}

We test the trained models on four hypernymy data sets for nouns, namely Kotlerman2010 \citep{kotlerman2010balapinc}, LEDS \citep{baroni-etal-2012-entailment}, WBLESS \citep{weeds-etal-2014-learning}, and EVALution \citep{santus-etal-2015-evalution}.
Each of them consists of a set of word pairs, each with a label indicating whether the second word is a hypernym of the first word.
We removed the out-of-vocabulary instances from all data sets, and non-nouns from EVALution during the evaluation.
\autoref{tab:test_sets} reports the statistics of the test sets data.
We report the AUC as in \autoref{sec:exp_synth}.

\begin{table}[th]
  \small
  \centering
  \begin{threeparttable}
    \begin{tabular}{lcc}
    \toprule
    \textbf{Test Set} & \textbf{\# Positive} & \textbf{\# Negative} \\
    \midrule
Kotlerman2010 & 880 [831] & 2058 [1919]  \\
LEDS & 1385 [1344] & 1385 [1342]  \\
WBLESS & 834 [830] & 834 [813]  \\
Evalution & 1592 [1352] & 4561 [3241]  \\
    \bottomrule  
    \end{tabular}
  \end{threeparttable}
  \caption{Class distributions of test sets. In brackets are the numbers after removal of OOV instances and non-nouns.}
  \label{tab:test_sets}
\end{table}

In addition, we use WBLESS for further performance analysis, which provides categorizations of the negative instances.
Each of the negative instances is either a hyponymy pair, co-hyponymy pair, meronymy pair, or random pair.



\subsection{Baselines}
Following \citet{roller-etal-2018-hearst}, we implement five distributional methods and train them on Wikiwoods using the distributional space described in \autoref{subsec:synth_eval1}.
Apart from the two DIH measures in \autoref{subsec:synth_eval1}, we use SLQS \citep{santus-etal-2014-chasing}, a word generality measure that rests on another hypothesis that general words mostly appear in uninformative contexts:
\begin{equation}
    \begin{aligned}
        \textup{SLQS}(r_1, r_2) & = 1 - \frac{E_{r_1}}{E_{r_2}} \\
        \textup{where}~ E_{r_i} & = \textup{median}^N_{j=1} [H(c_j)]
        \nonumber
    \end{aligned}
\end{equation}
For each word $r_i$, the median of the entropies of $N$ most associated contexts (as measured by local mutual information) is computed, where
$H(c_j)$ denotes the Shannon entropy of the associated context $c_j$.
Then, SLQS compares the generality of two words by their medians.
$N$ is chosen to be 50 following \citet{santus-etal-2014-chasing}.
We also include cosine similarity (Cosine) of $u^{(r_1)}$ and $u^{(r_2)}$, and SLQS--Cos, which multiplies the SLQS measure by Cosine since the SLQS measure only considers generality but not similarity.

\subsection{Results}

\begin{table}[th]
  \small
  \centering
  \setlength{\tabcolsep}{0.1cm} 
  \begin{threeparttable}
    \begin{tabular}{lcccc}
    \toprule
    \textbf{Model} & Kotlerman2010 & LEDS & WBLESS & Evalution \\
    \midrule
Cosine & .701 & .782 & .620 & .526 \\
WeedsPrec & .674 & .897 & .709 & .650 \\
invCL & .679 & .905 & .707 & .620 \\
SLQS & .491 & .480 & .568 & .532 \\
SLQS--Cos & .489 & .477 & .557 & .532 \\
\textsc{Fds} & .473 & .650 & .508 & .459 \\
\textsc{Fds}\textsubscript{$\forall$/2} & .558 & .759 & .660 & .583 \\
\textsc{Fds}\textsubscript{$\forall$} & .550 & .735 & .655 & .554 \\
    \bottomrule  
    \end{tabular}
  \end{threeparttable}
  \caption{AUC on the test sets.}
  \label{tab:real_AUC}
\end{table}

\autoref{tab:real_AUC} shows the results on the four test data sets.
The DIH baselines are competitive and nearly outperform all models across the test sets.
\textsc{Fds}\textsubscript{$\forall$} and \textsc{Fds}\textsubscript{$\forall$/2} both outperform \textsc{Fds} considerably across the test sets.
This reflects that including the proposed $\forall$-objective in training is useful for extracting hypernymy information in a corpus.
Compared to the 2.7-billion-token corpus used by \citet{santus-etal-2014-chasing} in training SLQS, we suggest that the Wikiwoods corpus is too small for SLQS to obtain meaningful contexts of the median entropy:
setting $N$ to be small results in frequent contexts that are not representative of the nouns, whilst setting it large would require a disproportionate number of contexts for the infrequent words.

\begin{table}[th]
  \small
  \centering
  \setlength{\tabcolsep}{0.05cm} 
  \begin{threeparttable}
    \begin{tabular}{lcccc}
    \toprule
    \textbf{Model} & Hyponymy & Co-hyponymy & Meronymy & Random \\
    \midrule
Cosine & .511 & .369 & .683 & .924 \\
WeedsPrec & .754 & .615 & .631 & .843 \\
invCL & .745 & .568 & .652 & .872 \\
SLQS & .606 & .551 & .590 & .524 \\
SLQS--Cos & .581 & .525 & .574 & .547 \\
\textsc{Fds} & .596 & .288 & .561 & .587 \\
\textsc{Fds}\textsubscript{$\forall$/2} & .783 & .612 & .549 & .704 \\
\textsc{Fds}\textsubscript{$\forall$} & .783 & .625 & .527 & .691 \\
    \bottomrule  
    \end{tabular}
  \end{threeparttable}
  \caption{AUC on the sub-categories of WBLESS.}
  \label{tab:wbless}
\end{table}

\autoref{tab:wbless} shows the results on the WBLESS sub-categories.
It is shown that \textsc{Fds}\textsubscript{$\forall$} is stronger than the DIH baselines in distinguishing between hyponymy and hypernymy pairs, and between co-hyponymy and hypernymy pairs, while weaker for meronymy or random pairs.
\textsc{Fds}\textsubscript{$\forall$} and \textsc{Fds}\textsubscript{$\forall$/2} outperform \textsc{Fds} in three out of the four sub-categories, with much higher distinguishing power for co-hyponymy and hyponymy.
These imply that the $\forall$-objective makes FDS more sensitive to the relative generality than the similarity of word pairs.

\section{Conclusion}
\label{sec:conc}

We have discussed how Functional Distributional Semantics (FDS) can provide a truth-conditional representation for hypernymy and demonstrate that it is learnable from the distributional information in a corpus.
On synthetic data sets, we confirm that FDS learns hypernymy under the Distributional Inclusion Hypothesis (DIH), and under the reverse of the DIH if the proposed objective for universal quantifications is applied.
On real data sets, the proposed objective substantially improves FDS performance on hypernymy detection.
We hope that this work provides insights into FDS models and hypernymy learning from corpora in general.

\section*{Limitations}
\label{limiations}
The proposed representation of hypernymy in FDS compares the semantic functions of DMRS predicate pairs.
Following previous implementations of Functional Distributional Semantics, a semantic function is a linear classifier.
Consequently, each DMRS predicate is assumed to have only one sense.
Modelling polysemy would require more expressive parametrizations of semantic functions, which can pose additional challenges to model training, and the hypernymy representation would possibly need to be revised.
Such an approach is considered out of the scope of this work.

\section*{Ethics Statement}

We anticipate no ethical issues directly stemming from our experiments.


\bibliography{anthology, custom, guy-thesis, pixie}
\bibliographystyle{acl_natbib}

\appendix

\section{Derivation of Hypernymy Conditions}
\label{app:derivation_hyp_def}
Consider \eqref{eq:fds_hyp_def}. $\forall z$:
\begin{align*}
    t^{(r_H, 0)}(z) & > t^{(r_h, 0)}(z) \nonumber \\
    S \!\left( {v^{(r_H, 0)}}^\top \! z + b^{(r_H, 0)} \right) & > S \!\left( {v^{(r_h, 0)}}^\top \! z + b^{(r_h, 0)} \right) \nonumber \\
    \shortintertext{$S$ is monotonic, so $\forall z$:}
    {v^{(r_H, 0)}}^\top z + b^{(r_H, 0)} & > {v^{(r_h, 0)}}^\top z + b^{(r_h, 0)} \nonumber \\
    b^{(r_H, 0)} - b^{(r_h, 0)} & >  \left (v^{(r_h, 0)} - v^{(r_H, 0)} \right )^\top z \nonumber
\end{align*}

For $z$ on a unit hypercube, this is equivalent to
\begin{align}
    b^{(r_H, 0)} - b^{(r_h, 0)} & > \max_{\lVert z \rVert_\infty = 1} \left (v^{(r_h, 0)} - v^{(r_H, 0)} \right )^\top z \nonumber
\end{align}
Note that $\argmax_{z_i} \left (v^{(r_h, 0)} - v^{(r_H, 0)} \right )^\top z = \textrm{sgn}({v^{(r_h, 0)}_i - v^{(r_H, 0)}_i})$ where $\textrm{sgn}$ is the sign function. Hence, we have
\begin{equation}
    b^{(r_H, 0)} - b^{(r_h, 0)} > {\left \lVert v^{(r_H, 0)} - v^{(r_h, 0)} \right \rVert}_1 \nonumber
\end{equation}

For $z$ on a unit hypersphere, this is equivalent to
\begin{align*}
    \label{eq:app_sphere}
    b^{(r_H, 0)} - b^{(r_h, 0)} & > \max_{\lVert z \rVert_2 = 1} \left (v^{(r_h, 0)} - v^{(r_H, 0)} \right )^\top z
\end{align*}
Note that $\argmax_{z} \left (v^{(r_h, 0)} - v^{(r_H, 0)} \right )^\top z  = \frac{v^{(r_h, 0)} - v^{(r_H, 0)}}{\left\lVert v^{(r_h, 0)} - v^{(r_H, 0)} \right\rVert_2}$.
Hence, we have
\begin{equation}
    b^{(r_H, 0)} - b^{(r_h, 0)} > {\left \lVert v^{(r_H, 0)} - v^{(r_h, 0)} \right \rVert}_2 \nonumber
\end{equation}

\section{Derivation of Transitivity}
\label{app:derivation_trans}
\begin{align*}
    & s(r_1, r_2) + s(r_2, r_3) \\
    & = b^{(r_2, 0)} - b^{(r_1, 0)} - {\left \lVert v^{(r_2, 0)} - v^{(r_1, 0)} \right \rVert}_p \\
    & \phantom{\;=\;} + b^{(r_3, 0)} - b^{(r_2, 0)} - {\left \lVert v^{(r_3, 0)} - v^{(r_2, 0)} \right \rVert}_p \\
    & = b^{(r_3, 0)} - b^{(r_1, 0)} - \\
    & \phantom{\;=\;} \left ({\left \lVert v^{(r_2, 0)} - v^{(r_1, 0)} \right \rVert}_p + {\left \lVert v^{(r_3, 0)} - v^{(r_2, 0)} \right \rVert}_p \right )
\end{align*}
By the Minkowski inequality, the last term is greater than ${\left \lVert v^{(r_3, 0)} - v^{(r_1, 0)} \right \rVert}_p$. Besides, when $s(r_1, r_2) > 0$ and $s(r_2, r_3) > 0$, $s(r_1, r_2) + s(r_2, r_3) > 0$. Hence,
\begin{align*}
    b^{(r_3, 0)} - b^{(r_1, 0)} - {\left \lVert v^{(r_3, 0)} - v^{(r_1, 0)} \right \rVert}_p & > 0 \\
    s(r_3, r_1) & > 0
\end{align*}

\section{Intuition behind Hypernymy Learning by FDS under the DIH}
\label{app:intuition}

We hypothesize that the way that FDS models are trained allows hypernymy learning under the DIH.
During training described in \autoref{subsec:model_train}, the approximate posterior distributions of pixies are first inferred from the observed graph.
After variational inference, the semantic functions of the observed predicates are optimized to be true of the inferred pixie distributions.
This process is analogous to the following process under a model-theoretic approach: the entities described by a sentence are first identified, and then the truth conditions of predicates over the entities are updated as asserted by the sentence.

Under the DIH, the contexts of nouns are also contexts of their hypernyms.
The local predicate--argument information of nouns, i.e. contexts, is thus repeated for their hypernyms for inference during training.
Consequently, the semantic functions of hypernyms are trained to return values at least as high as those of their hyponyms over the pixie distributions inferred from the same contexts.
The additional contexts appearing exclusively with the hypernyms will further increase the probability of truths of the hypernyms over the pixie space.
By \eqref{eq:fds_hyp_def}, hypernymy should thus be learnt under the DIH.

\section{Training Details}
\label{app:train}

\subsection{Hyperparameters and Tuning}
\label{app:hyperparam}

For all the experiments, the hyperparameters of the FDS models largely follow that of FDSAS\textsubscript{id} in \citet{lo-etal-2023-functional} except that we set $\beta_1$ to 0.5 instead of 0.
The consequence is that the inferred pixie distributions during VAE training will be centred closer to the origin.
This is motivated by our decision in \autoref{subsec:rep_hyp} that pixies are only meaningful within the unit hypersphere or hypercube.

Here are the changes exclusive to the experiments on the synthetic data sets.
We set $K$ to 1 and perform random negative sampling without weighing by unigram distribution, which trains models maximally using information from the data with minimal assumptions needed for the negative samples.
We set the learning rate to 0.01.
For experiments on $H_\textup{chains}$, $d$ is set to 2. For $H_\textup{WN}$, $d$ is set to 50. For the remaining topologies, $d$ is set to 10.
The models are trained for 2 epochs for $H_\textup{WN}$, and 5000 epochs for the rest.

\subsection{Computational Configurations}
\label{app:comp_config}
All models are implemented in PyTorch \citep{NEURIPS2019_bdbca288} and trained with distributed data parallelism on three NVIDIA GeForce GTX 1080 Ti. Training a run of \textsc{Fds} or \textsc{Fds}\textsubscript{$\forall$} on Wikiwoods takes about 360 GPU hours.

\section{Visualization of Semantic Functions}
\label{app:vis}

\begin{figure}[t]

    \centering
    \small
    \begin{subfigure}[t]{0.49\linewidth}
        \centering
        \includegraphics[width=\linewidth]{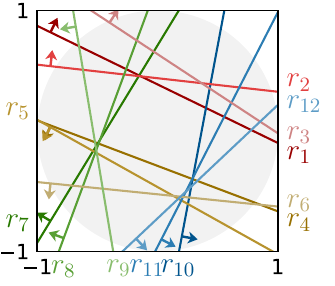}
        \caption{\textsc{Fds} on DIH corpus}
        \label{subfig:fds_dih}
    \end{subfigure}
    \hfill
    \begin{subfigure}[t]{0.49\linewidth}
        \centering
        \includegraphics[width=\linewidth]{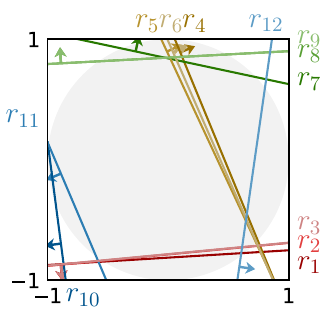}
        \caption{\textsc{Fds}\textsubscript{$\forall$} on DIH corpus}
        \label{subfig:fdsa_dih}
    \end{subfigure}
    \par\bigskip
    \begin{subfigure}[t]{0.49\linewidth}
        \centering
        \includegraphics[width=\linewidth]{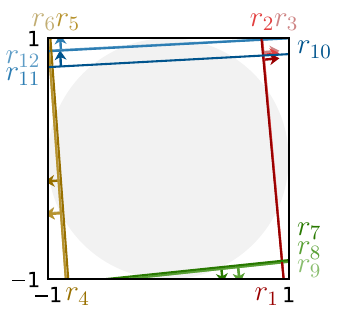}
        \caption{\textsc{Fds}\textsubscript{$\forall$} on rDIH corpus}
        \label{subfig:fdsa_rdih}
    \end{subfigure}
    \hfill
    \begin{subfigure}[t]{0.49\linewidth}
        \centering
        \includegraphics[width=\linewidth]{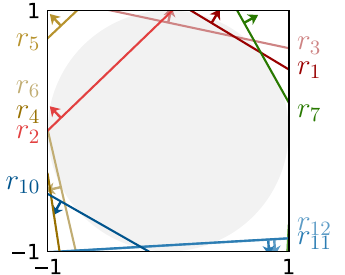}
        \caption{\textsc{Fds} on rDIH corpus}
        \label{subfig:fds_rdih}
    \end{subfigure}
    \caption{Visualization of semantic functions of a run trained on $H_\textup{chains}$. Each plot shows a pixie space in a unit square (unit circle in grey). Each line plots $t^{(r_i, 0)}(z) = 0$ and the arrow points to the pixie subspace where $t^{(r_i, 0)}(z) > 0$.}
    \label{fig:4_chains}
\end{figure}

A visualization of results on $H_\textup{chains}$ is provided in \autoref{fig:4_chains}.
As seen in \Autorefs{subfig:fds_dih} and \ref{subfig:fdsa_rdih},
training \textsc{Fds} on the DIH corpus and \textsc{Fds}\textsubscript{$\forall$} on the rDIH corpus both result in four nicely divided pixie subspaces, each for one of the four hypernymy chains, as shown in the plots on the left column.
In contrast, applying the other models sometimes gives badly learnt semantic functions, as shown in \Autorefs{subfig:fdsa_dih} and \ref{subfig:fds_rdih}.
For example, $t^{(r_{12}, 0)}$ points to the opposite direction of $t^{(r_{10}, 0)}$ and $t^{(r_{11}, 0)}$ in \autoref{subfig:fdsa_dih}.

\end{document}